\documentclass[conference]{IEEEtran}
\IEEEoverridecommandlockouts
\usepackage{cite}
\usepackage{amsmath,amssymb,amsfonts}
\usepackage{algorithm}
\usepackage{algorithmic}
\usepackage{graphicx}
\usepackage{textcomp}
\usepackage{xcolor}
\usepackage{booktabs}
\usepackage{url}
\usepackage{tikz}
\usetikzlibrary{positioning,arrows.meta,fit,calc}
\def\BibTeX{{\rm B\kern-.05em{\sc i\kern-.025em b}\kern-.08em
    T\kern-.1667em\lower.7ex\hbox{E}\kern-.125emX}}

\newtheorem{definition}{Definition}

\begin{document}

\title{Self-Crystallizing Behavior Trees: Amortizing Foundation-Model Reasoning into Deterministic Control for Autonomous Drones}

\author{\IEEEauthorblockN{Behnam Moradi}
\IEEEauthorblockA{\textit{Software Department, KavAI.}\\
California, USA \\
behnam.moradi@kavai.com}
}

\maketitle

\begin{abstract}
Classical robot control architectures, Finite State Machines (FSMs) and Behavior Trees (BTs), are deterministic, inspectable, and fast, but fixed at design time: they fail on the long tail of situations their designers did not anticipate. Foundation models can synthesize new control logic at runtime, at the price of latency, cost, and non-determinism that are hard to accept in safety-critical aerial systems. Existing LLM-driven approaches keep the model permanently in the loop and discard what it produces after each use. We propose \emph{Self-Crystallizing Behavior Trees} (SCBT), in which the foundation model is a \emph{compiler} rather than a \emph{runtime}. When a node of a vehicle's BT fails, an LLM proposes a minimal subtree \emph{patch} in a constrained, mission-agnostic grammar; CMA-ES tunes the patch's continuous parameters in simulation; a contextual bandit decides whether a stored patch or a new LLM call is worth paying for; and a three-stage admission pipeline (static check, invariant monitors over randomized replays, and an empirical margin over the unpatched tree) admits only patches that demonstrably recover. Admitted patches are stored in a fleet library indexed by failure context and, once the LLM call rate has decayed, \emph{crystallized} into a conventional deterministic BT that no human wrote. We implement SCBT on Microsoft Project AirSim with a schema-constrained LLM pipeline (Google ADK, Gemini~3.1~Pro) and a lockstep simulator. Over three seeds of a 20-episode learning phase with injected wind-gust and battery failures, SCBT used $11\pm1.4$ LLM calls against 26 for whole-tree regeneration and 22 for patching without a library, with 0.33 calls per episode in the second half. The crystallized trees, with 28 and 33 nodes and two patches each, reached $97\pm2\%$ mission success on the curriculum with no learning component running, against 12\% for the unpatched base tree and 98\% for a hand-coded expert tree; re-ticked offline against recorded observations they reproduce their online decision traces exactly. Two negative results are reported: replay admission as parameterized rejected useful wind recoveries because it attributed wind-induced geofence excursions to the patch, and the static-only ablation generalized better to held-out failures as a consequence.
\end{abstract}


\section{Introduction}
\label{sec:intro}

For half a century robot behavior has been specified explicitly, first as FSMs and, over the last decade, as BTs~\cite{colledanchise2018bt,iovino2022survey}. These architectures are attractive precisely because they are \emph{closed}: every branch can be inspected, every transition is deterministic, and the controller runs at a fixed rate on embedded hardware. The same closedness is their limitation. A BT cannot execute a branch that was never written, and in open environments the set of situations that require a branch is effectively unbounded. We call this \emph{design-time rigidity}.

Foundation models offer an apparent escape: LLMs write robot programs from natural-language goals~\cite{liang2023cap}, generate BT XML on demand~\cite{izzo2024btgenbot,zhou2024llmbt}, ground plans in perceptual affordances~\cite{ahn2022saycan,huang2023voxposer}, and coordinate heterogeneous teams~\cite{mandi2024roco,chen2024scalable}. But these systems share a structural assumption: the model stays in the control loop for the lifetime of the system, and what it produces is discarded after use. For aerial robots this has three consequences. \textbf{Latency}: in our implementation a single structured proposal from a frontier model takes 31--105\,s (two schema-constrained calls); a vehicle in a wind gust cannot wait for it. \textbf{Cost}: operating cost grows with activity. \textbf{Non-determinism}: two flights under identical conditions may execute different logic, which defeats the certification argument that makes BTs acceptable in the first place.

The two camps are usually framed as a trade-off between adaptability and determinism. We argue the trade-off is an artifact of \emph{when} the foundation model is invoked. If the model is called only when the existing control structure fails, if what it produces is verified and \emph{retained}, and if a learned policy stops calling it once stored structure suffices, then adaptability is paid for once, during a learning phase, and the deployed system is an ordinary deterministic BT.

\subsection{Approach}

SCBT assigns each mechanism to the job it is suited for:
\begin{itemize}
\item The \textbf{LLM proposes structure}: given a failed node and its context, it emits one minimal, local, mission-agnostic patch in a constrained grammar. Structure synthesis is discrete, symbolic, and rare.
\item \textbf{CMA-ES tunes parameters}: each patch declares its uncertain numeric ports as tunables with bounds; these are optimized against simulated replays before admission.
\item A \textbf{contextual bandit selects}: among stored patches that match the failure context and the option of a new LLM call, with the call's cost in the reward.
\item A \textbf{three-stage admission pipeline} gates library entry: static check against the grammar and the vehicle's capability manifest; invariant monitors over randomized replays; and an empirical success margin over the tree as it currently stands.
\item A \textbf{deterministic BT executes}: the vehicle controller is always a plain BT ticked in lockstep with the simulator. No neural component is in the tick path.
\end{itemize}

\subsection{Contributions}
\begin{enumerate}
\item The SCBT framework: failure interception, a mission-agnostic patch grammar with anchor-port references, a context-indexed fleet library, bandit-based selection with an explicit LLM-call arm, and replay-based admission with CMA-ES tuning (Sec.~\ref{sec:framework}).
\item An implementation on Project AirSim with a schema-constrained ADK agent pipeline, a lockstep tick loop at $10\times$ real time whose decision traces are an exact function of the observation sequence, and a failure-injection curriculum calibrated against the simulator's measured response (Sec.~\ref{sec:impl}).
\item An evaluation against hand-coded, regenerate-on-failure, and five ablated baselines over a frozen phase in which all learning is disabled, with an exact decision-determinism test, and an analysis of what the admission pipeline caught and what it wrongly rejected (Secs.~\ref{sec:exp},~\ref{sec:results}).
\end{enumerate}

\section{Related Work}
\label{sec:related}

\subsection{Classical Control Architectures}
FSMs require transitions from every state to handle global events, so transition count grows quadratically with state count~\cite{colledanchise2018bt}. BTs~\cite{ogren2012uav,marzinotto2014unified} standardize the node interface to a ternary status and decouple nodes; the survey by Iovino et al.~\cite{iovino2022survey} documents the gains in modularity and reactivity and the persistent limitation that BT structure is fixed at design time. Learned BTs via genetic programming~\cite{colledanchise2019learning,iovino2021gp} or learned leaf policies~\cite{pereira2015framework} learn from reward alone and are sample-hungry when the needed structural change is semantically obvious.

\subsection{Foundation Models for Control Synthesis}
Code as Policies~\cite{liang2023cap} prompts an LLM with a robot API to write programs; SayCan~\cite{ahn2022saycan} and Inner Monologue~\cite{huang2022inner} ground plans in affordances and feedback; VoxPoser~\cite{huang2023voxposer} composes LLM reasoning and VLM grounding into value maps; VLA models~\cite{brohan2023rt2,kim2024openvla} map pixels to actions. Closest to us are BTGenBot~\cite{izzo2024btgenbot}, which fine-tunes small LLMs to emit BehaviorTree.CPP XML and validates it syntactically and in simulation, and LLM-BT~\cite{zhou2024llmbt}. These regenerate structure on demand and do not retain, verify against an empirical margin, or learn when to stop calling the model. Surveys~\cite{firoozi2024foundation,xu2024foundationmodels} discuss the latency, cost, and hallucination risks that motivate our design.

\subsection{Combining LLMs with Learning}
Eureka~\cite{ma2024eureka} has an LLM write reward functions that RL optimizes; RL-GPT~\cite{liu2024rlgpt} lets an LLM decide what to code and what to learn; Voyager~\cite{wang2023voyager} keeps a growing library of LLM-written skills retrieved by embedding. SCBT shares the library with Voyager and the LLM/optimizer division of labor with Eureka, but the stored artifacts are BT subtrees, admission is gated by simulated invariants and an empirical margin, and the stated end state is a system with no model running.

\subsection{Multi-Robot Coordination and Verification}
Classical MRTA~\cite{gerkey2004taxonomy,kuhn1955hungarian,gerkey2002sold,nunes2015tessi} requires pre-defined tasks with comparable costs. LLM coordination~\cite{mandi2024roco,chen2024scalable} describes robots by textual capability profiles; we adopt that idea as a \emph{capability manifest} that gates which leaves a patch may use and declares substitutions across vehicle classes. Prior verification of generated control checks syntax and simulated execution~\cite{izzo2024btgenbot}; we add invariant monitors, a landing-reserve constraint, and a required margin over the unpatched tree.

\section{Problem Formulation}
\label{sec:problem}

\subsection{Behavior Trees}
A BT is a rooted tree whose internal nodes are control-flow nodes (Sequence, Fallback, Parallel, decorators) and whose leaves are actions and conditions~\cite{colledanchise2018bt}. It is ticked at a fixed rate; each node returns Success, Failure, or Running. We use py\_trees~\cite{pytrees} with an XML dialect that mirrors BehaviorTree.CPP: Sequence/Fallback with memory, Reactive variants that re-evaluate all children every tick, Parallel, Inverter, ForceSuccess/Failure, Retry, Repeat. Leaves declare typed \emph{ports}; numeric ports may carry bounds, and bounded ports are \emph{tunable}. We write $\theta$ for a vector of tunable port values.

\subsection{Vehicles and Capability Manifests}
Each vehicle has a class $m$ with a manifest $C_m$: the leaf types the class may execute, with port signatures and one-line descriptions, plus \emph{substitutions} (e.g.\ a camera-less class substitutes \texttt{RequestScout} for \texttt{InspectRegion}). Classes differ physically (frame mass, speed cap enforced by the leaves, battery drain).

\subsection{Missions and Success}
A mission plan $(T_m,\ G_{\mathrm{req}},\ G_{\mathrm{opt}})$ is a base tree plus required and optional \emph{goals}; goal leaves record completion on the blackboard. A mission \emph{succeeds} if all required goals are reached and the vehicle lands. Mission \emph{value} weights required goals 1 and optional goals 0.5. An abort to the hand-coded safe branch after the required goals are met is not a failure; an abort before them is. This definition matters: a recovery that merely returns to launch scores zero when a required goal is outstanding, which is what makes the battery curriculum (Sec.~\ref{sec:curriculum}) a prioritization problem rather than a trivial one.

\subsection{Failure Events}
\begin{definition}[Failure event]
A failure event on a vehicle is $e=(v_f, c)$ where $v_f$ is the deepest leaf that returned Failure in the tick in which the mission-critical subtree failed, and $c=(s, b, m, \tau, X)$ with $s$ a textual context, $b$ the telemetry snapshot, $m$ the class, $\tau$ the last $k$ status changes in the enclosing subtree, and $X$ the XML of the enclosing subtree with $v_f$ marked.
\end{definition}
In the headless setting $s$ is generated from telemetry (failed node type, leaf feedback, altitude, speed, wind, battery, GPS and link status); on a rendering host it is a VLM description. The vehicle class is deliberately excluded from $s$ so that retrieval surfaces cross-class candidates.

\subsection{Objective}
Over a sequence of failure events the learning phase maximizes
\begin{equation}
\sum_t \big(y_t - \kappa\,u_t - \lambda\,\Delta t_t/100 - \mu\,\Delta E_t/10\big),
\label{eq:objective}
\end{equation}
where $y_t$ is mission success after event $t$, $u_t$ indicates an LLM call, $\Delta t_t$ (s) and $\Delta E_t$ (\% battery) are the time and energy from the event to the end of the episode, and $\kappa,\lambda,\mu$ weight cost, time and energy. The deployment objective is that the crystallized tree matches learning-phase success with no LLM, optimizer, or bandit running.

\section{The SCBT Framework}
\label{sec:framework}
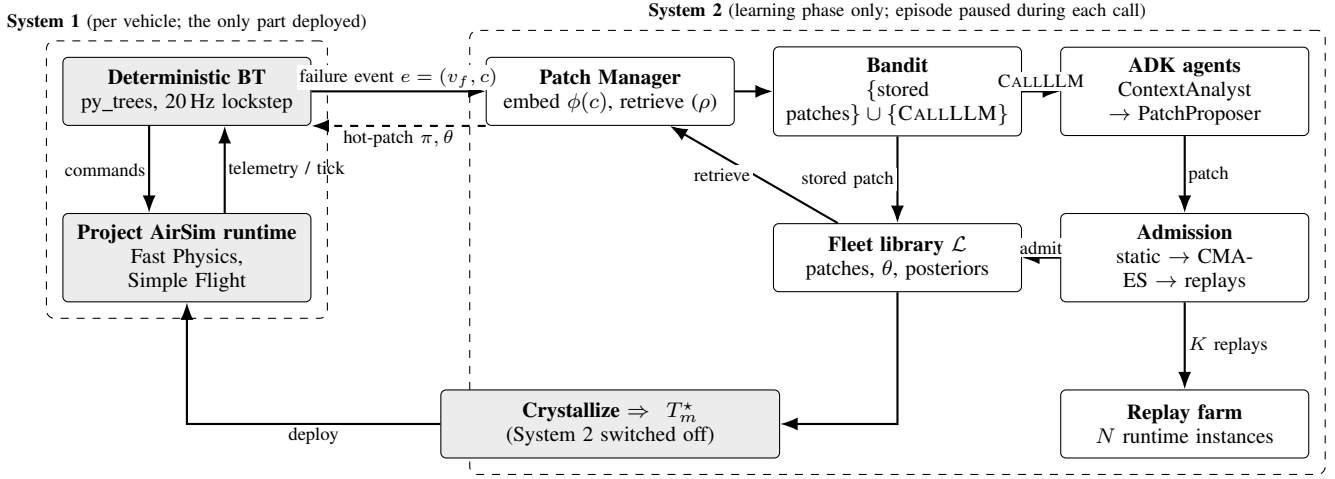
\begin{figure*}[t]
\centering
\begin{tikzpicture}[
  font=\footnotesize, >=Latex, x=1cm, y=1cm,
  box/.style={draw, rounded corners=2pt, align=center, inner sep=4pt, minimum height=9mm, text width=3.0cm},
  arr/.style={->, thick}, lab/.style={font=\scriptsize, fill=white, inner sep=1pt}]
\node[box, fill=gray!15] (bt)  at (0,2.2) {\textbf{Deterministic BT}\\py\_trees, 20\,Hz lockstep};
\node[box, fill=gray!15] (sim) at (0,0)   {\textbf{Project AirSim runtime}\\Fast Physics, Simple Flight};
\draw[arr] ([xshift=-5mm]bt.south) -- node[lab, left]{commands} ([xshift=-5mm]sim.north);
\draw[arr] ([xshift=5mm]sim.north) -- node[lab, right]{telemetry / tick} ([xshift=5mm]bt.south);
\node[draw, dashed, rounded corners=3pt, inner sep=6pt, fit=(bt)(sim)] (s1) {};
\node[font=\scriptsize, anchor=south] at (s1.north) {\textbf{System 1} (per vehicle; the only part deployed)};
\node[box] (mgr)    at (5.6,2.2) {\textbf{Patch Manager}\\embed $\phi(c)$, retrieve ($\rho$)};
\node[box] (bandit) at (9.4,2.2) {\textbf{Bandit}\\$\{$stored patches$\}\cup\{\textsc{CallLLM}\}$};
\node[box] (llm)    at (13.2,2.2) {\textbf{ADK agents}\\ContextAnalyst $\to$ PatchProposer};
\node[box] (adm)    at (13.2,0)  {\textbf{Admission}\\static $\to$ CMA-ES $\to$ replays};
\node[box] (farm)   at (13.2,-2.2) {\textbf{Replay farm}\\$N$ runtime instances};
\node[box] (lib)    at (9.4,0)   {\textbf{Fleet library} $\mathcal{L}$\\patches, $\theta$, posteriors};
\draw[arr] (mgr) -- (bandit);
\draw[arr] (bandit) -- node[lab, above]{\textsc{CallLLM}} (llm);
\draw[arr] (llm) -- node[lab, right]{patch} (adm);
\draw[arr] (adm) -- node[lab, right]{$K$ replays} (farm);
\draw[arr] (adm) -- node[lab, above]{admit} (lib);
\draw[arr] (lib) -- node[lab, left]{retrieve} (mgr);
\draw[arr] (bandit) -- node[lab, left]{stored patch} (lib);
\node[draw, dashed, rounded corners=3pt, inner sep=6pt, fit=(mgr)(bandit)(llm)(adm)(farm)(lib)] (s2) {};
\node[font=\scriptsize, anchor=south] at (s2.north) {\textbf{System 2} (learning phase only; episode paused during each call)};
\draw[arr] (bt.east) -- node[lab, above]{failure event $e=(v_f,c)$} (mgr.west);
\draw[arr, dashed] (mgr.south west) -- node[lab, below, sloped]{hot-patch $\pi,\theta$} (bt.south east);
\node[box, fill=gray!15, text width=4.2cm] (cry) at (5.6,-2.2) {\textbf{Crystallize} $\Rightarrow T^\star_m$\\(System 2 switched off)};
\draw[arr] (lib.south) |- (cry.east);
\draw[arr] (cry.west) -| node[lab, below, pos=0.25]{deploy} (sim.south);
\end{tikzpicture}
\caption{SCBT architecture. The vehicle's tick path contains only a deterministic BT. The Patch Manager runs above it during the learning phase, with the live episode paused during each call, and is absent after crystallization.}
\label{fig:arch}
\end{figure*}

Fig.~\ref{fig:arch} shows the components and the boundary between what is deployed and what only runs during learning.

\subsection{Failure Interception}
A \texttt{MissionCritical} decorator wraps the mission subtree. When its child returns Failure and patching is enabled, it returns Running for that tick and raises a flag; the runner pauses the simulator, constructs $e$, and consults the Patch Manager. If a patch is applied, the mission subtree restarts on the next tick (leaves are idempotent: takeoff when airborne and navigation to a reached waypoint return Success immediately). If no patch is available, the decorator returns Failure on the next tick \emph{without ticking its child}, and the root Fallback proceeds to the hand-coded abort branch. With patching disabled (frozen phase, baselines) the decorator is a pure pass-through. Each episode has a budget of $P_{\max}$ patches and $F_{\max}$ failures of the same node; already-applied patches are excluded from fallback selection. Without these limits the manager re-applied the same patch on the same anchor until the tree exceeded the interpreter's recursion limit.

\subsection{Patch Grammar}
\label{sec:patches}
\begin{definition}[Patch]
$\pi=(a, o, \sigma, \Theta_\sigma, r)$: anchor node name $a$, operator $o\in\mathcal{O}=\{$\textsc{WrapFallback}, \textsc{GuardWith}, \textsc{WrapRetry}, \textsc{InsertBefore}, \textsc{InsertAfter}, \textsc{ReplaceSubtree}$\}$, subtree $\sigma$ whose leaves are in $C_m$ (or have substitutions), tunable box $\Theta_\sigma$ with per-port bounds and initial values, and a rationale $r$.
\end{definition}
\textsc{WrapFallback} replaces $a$ by $\mathrm{Fallback}(a,\sigma)$; \textsc{GuardWith} by $\mathrm{Sequence}(\sigma,a)$; \textsc{ReplaceSubtree} is the only operator that deletes logic and requires a rationale. Two features keep patches \emph{mission-agnostic}: an attribute value \texttt{\{@p\}} refers to port $p$ of the anchor leaf and is resolved when the patch is applied, and a \texttt{GoToRequiredGoal} leaf flies to the next outstanding required goal read from the blackboard. Literal coordinates copied from the event are therefore unnecessary; during development, a patch that contained them was reused on a different mission and flew to the wrong place, which is why the rule exists. Wrapper nodes created by a patch are named after the patch identifier and nodes in $\sigma$ whose names collide with existing ones are renamed, so anchors remain unique and a crystallized tree is readable.

\subsection{LLM Proposer}
The proposer is a Google ADK~\cite{adk} \texttt{SequentialAgent} of two \texttt{LlmAgent}s with Pydantic output schemas: a \emph{ContextAnalyst} producing a diagnosis (summary, likely cause, strategy, anchor) and a \emph{PatchProposer} producing exactly one patch. Malformed output is rejected by the schema before our checker runs. The message contains the rendered event, the mission's goal structure, the success criterion and the hard invariants, the manifest, the grammar, and up to three nearest library entries \emph{with their outcome records} (admission status, uses, successes, rejection reason). A \emph{repair loop} re-prompts with the static checker's message up to $R$ times. The success criterion was added after early runs showed the model proposing a safe return-to-launch that abandoned the required goal; stating the objective explicitly is a prompt-engineering fix, but the pipeline's protection against such patches is admission, not prompting.

\subsection{Embedding, Library, and Bandit}
\label{sec:bandit}
$\phi(c)$ concatenates a MiniLM sentence embedding~\cite{reimers2019sbert} of $s$, one-hot encodings of the failed node type and the class, and normalized telemetry. Retrieval uses cosine similarity on the text and node-type blocks only (class excluded) with threshold $\rho$. The library stores, per patch, the embedding, XML, tuned $\theta$ per class, admission status and reason, provenance (event, rationale, proposer, attempts, admission record), and use/success counts.

Selection is linear Thompson sampling~\cite{agrawal2013ts} over arms $\{\text{retrieved patches}\}\cup\{\textsc{CallLLM}\}$ with context $[\phi(c),1]$; new arms get an optimistic prior mean, \textsc{CallLLM} a neutral one. Rewards per (\ref{eq:objective}) are computed at episode end and credited to the arm played; when \textsc{CallLLM} produced a patch, that patch is additionally credited with the reward plus $\kappa$. As stored arms accumulate evidence, \textsc{CallLLM}, which always pays $\kappa$, is selected less often; LLM usage decays because it is learned, not scheduled. The frozen phase replaces sampling by the posterior-mean argmax.

\subsection{Replay-Based Admission and Tuning}
\label{sec:admission}
A \emph{replay} re-runs the originating episode (same mission, same failure type) on a dedicated simulator instance with a candidate tree compiled from the \emph{live tree as it currently stands} plus the patch, with patching disabled. Replaying from mission start, rather than respawning at the recorded state, keeps controller and sensor state consistent and costs one episode per replay (5--15\,s wall at $10\times$; four runtime instances serve replays in parallel). The event neighbourhood $\mathcal{N}(e)$ perturbs injection parameters by $\pm p$ and the trigger time by $\pm 2$\,s, \emph{clipped to the failure's catalogue range}; without clipping, a perturbed battery level drifted above the guard and the replay never failed.

\textbf{Tuning.} For $\dim\Theta_\sigma>0$, CMA-ES~\cite{hansen2001cmaes} in the unit box with budget $B$ maximizes
\begin{equation}
J(\theta) = \hat{p}_{\mathrm{succ}} - \lambda\,\bar{t}/100 - \mu\,\bar{E}/10 - \nu\,\mathbf{1}[\text{unsafe}],
\label{eq:J}
\end{equation}
where \emph{unsafe} means any invariant violation or landing reserve below 10\%. The safety term is in the objective because without it CMA-ES traded reserve for success rate and produced a patch that landed at 9.8\%.

\textbf{Stages.} (1) \emph{Static}: operator and anchor exist; leaves in $C_m$ or substitutable; ports exist, parse, and lie within bounds; placeholders declared and used; anchor references resolvable; depth limit; \textsc{ReplaceSubtree} has a rationale and does not remove \texttt{MissionCritical}. (2) \emph{Constraint}: $K$ replays of the tuned patch; per-tick monitors for geofence, altitude, speed (class cap with margin), inter-vehicle separation; reject on any violation tick or landing reserve $<10\%$. (3) \emph{Empirical}: success rate $\ge\rho_{\min}$ and $\ge$ the rate of the current tree without the candidate (same perturbation seeds) $+\delta$. The baseline is the \emph{current} tree, so a candidate for a failure the tree already handles is rejected for lack of margin; this was the most common stage-3 rejection in our runs. Admission errors reject rather than abort the run.

\subsection{Cross-Class Transfer}
A patch is a candidate for class $m'$ if every leaf is in $C_{m'}$ or has a substitution, which the static checker applies in place; $\theta$ is re-tuned from the originating class's optimum and the patch is re-admitted for $m'$. Retrieval is class-agnostic, so such candidates are visible.

\subsection{Crystallization}
\label{sec:crystal}
When the windowed LLM-call rate falls below $\epsilon$ and success is at least $\gamma$, each class's admitted patches are applied to $T_m$ in admission order with $\theta^{(m)}$ fixed; patches sharing an anchor become children of one Fallback ordered by bandit posterior mean. The result $T^\star_m$ runs with no LLM, optimizer, or bandit. Under the idealized assumptions of clustered contexts and a positive per-call admission probability, expected LLM calls are bounded by $Q|\mathcal{M}|/p$ plus bandit exploration; we state this only to make explicit what the empirical amortization curve measures (how far real contexts behave like a finite set of clusters at the chosen $\rho$).

\section{Implementation}
\label{sec:impl}

\textbf{Simulator.} Microsoft Project AirSim~\cite{projectairsim} (successor of AirSim~\cite{shah2017airsim}), headless runtime: Fast Physics, Simple Flight controller, IMU/GPS/barometer/battery sensors, no rendering and no mesh collisions. Scene and robot configurations are generated programmatically per class (frame mass, battery capacity and drain). The steppable clock is driven in \emph{lockstep}: the simulator is paused and advanced by exactly one tick period (50\,ms) per BT tick at $10\times$ real time. Two client-side races were removed so that the tick trace is a function of the observation sequence alone: commands issued during a tick are flushed, and a synchronous barrier request is issued, before the step; and action leaves decide completion from the per-tick telemetry snapshot rather than from the server's reply, whose arrival time is wall-clock dependent. The simulator itself is not bit-reproducible: its command-execution thread is wall-clock driven, and with no BT in the loop five identical lockstep runs produced three trajectories differing at the $10^{-4}$\,m level. End-to-end repetitions can therefore still differ by a threshold crossing; Sec.~\ref{sec:results} reports both the end-to-end and the exact offline measurement. Several runtime instances on distinct port pairs serve the live episode and the replay farm.

\textbf{BT runtime.} py\_trees~\cite{pytrees} with 17 leaf types over the Project AirSim Python client: \texttt{Takeoff}, \texttt{Land}, \texttt{ReturnToLaunch}, \texttt{GoToNED}, \texttt{GoToRequiredGoal}, \texttt{MoveRelative}, \texttt{Hover}, \texttt{WaitSimTime}, \texttt{SetSpeed}, \texttt{InspectRegion}/\texttt{RequestScout}, and conditions \texttt{IsAirborne}, \texttt{IsBatteryOk}, \texttt{IsWindOk}, \texttt{IsGpsOk}, \texttt{IsLinkOk}, \texttt{IsInsideGeofence}. Actions start an asynchronous simulator command on first tick and poll it; telemetry is refreshed once per tick into a dictionary that leaves read, so the tick trace is a function of the telemetry sequence. Blackboards are namespaced per (simulator instance, vehicle); the replay farm runs vehicles with the same names, and without the instance prefix the live tree commanded the farm's drone.

\textbf{Base trees.} Fig.~\ref{fig:basetree} shows the base tree, identical for all classes up to waypoints and limits. Missions: M1 \emph{transport} (deliver 40\,m north, required; return; land) and M2 \emph{survey} (three lanes of 30\,m; lane~1 required, lanes~2--3 optional; return; land). A nominal M1 uses 12.7\% battery and M2 20.7\%.

\begin{figure}[t]
\centering
\begin{minipage}{0.92\columnwidth}
\scriptsize\ttfamily
Fallback root\\
\hspace*{1em}MissionCritical mission\\
\hspace*{2em}Sequence mission\_seq\\
\hspace*{3em}Takeoff\\
\hspace*{3em}SetSpeed\\
\hspace*{3em}ReactiveSequence guarded\_flight\\
\hspace*{4em}IsBatteryOk min\_pct=25\\
\hspace*{4em}IsInsideGeofence\\
\hspace*{4em}Sequence waypoints\\
\hspace*{5em}GoToNED wp1 ... GoToNED wp\_n (goal\_id)\\
\hspace*{3em}Land\\
\hspace*{1em}Sequence abort\\
\hspace*{2em}ReturnToLaunch\\
\hspace*{2em}Land
\end{minipage}
\caption{Base tree $T_m$. Guards are re-evaluated every tick; a failure anywhere under \texttt{MissionCritical} raises a failure event, and if no patch is applied the hand-coded \texttt{abort} branch runs.}
\label{fig:basetree}
\end{figure}

\textbf{LLM and embedding.} Gemini~3.1~Pro (\texttt{gemini-3.1-pro-preview}) at temperature 0.2 through ADK; mean latency per structured call 31--105\,s depending on method (two calls per proposal). Sentence embedding \texttt{all-MiniLM-L6-v2}. A mock proposer that replays hand-written patches lets the full System-2 pipeline run offline for tests.

\textbf{Parameters.} $\rho=0.80$, $\kappa=0.30$, $\lambda=\mu=0.10$, $\nu=1$, $R=2$, $P_{\max}=3$, $F_{\max}=2$, $K=3$, $B=4$, $p=0.20$, $\rho_{\min}=0.75$, $\delta=0.25$.

\section{Experimental Design}
\label{sec:exp}

\subsection{Failure Curriculum}
\label{sec:curriculum}
Failures are injected at a random time in $[6,14]$\,s. Headless-capable: \textbf{F1} wind gust; \textbf{F2} GPS noise then loss; \textbf{F5} landing pad shifted; \textbf{F6} no-fly zone (geofence shrinks across the route); \textbf{F7} battery capacity reduced; \textbf{F8} link dropout; \textbf{F12} separation conflict (two vehicles). Obstacles, pad occupancy, camera dropout, and target search (F3, F4, F9--F11) require the rendering host and are future work. Two calibrations matter. \emph{F1}: the Fast Physics quadrotor rejects uniform wind up to 20\,m/s almost completely ($<1$\,m drift in 8\,s) and saturates above $\sim$25\,m/s (47\,m drift in 8\,s at 30\,m/s; a 40\,m leg under 25\,m/s ended 226\,m from its target); F1 therefore samples 22--28\,m/s for 15--30\,s, which is ``beyond tracking capability'' by measurement. \emph{F7}: battery set to 17--24\% (below the 25\% guard) with drain $\times 1.3$--$2.2$, so that completing the required leg under a reduced reserve is feasible in most but not all instances. Held-out variants use disjoint parameter sub-ranges.

\subsection{Baselines and Ablations}
\textbf{B0} hand-coded expert tree with recovery branches for all failures; \textbf{B1} whole-tree regeneration by the LLM on every failure, no library; \textbf{B2} patch grammar and admission, no library (every failure calls the LLM); \textbf{B3} no CMA-ES (LLM's initial values); \textbf{B4} no bandit (nearest neighbour, else LLM); \textbf{B5} static admission only; \textbf{B6} per-class libraries; \textbf{SCBT} full. B2--B6 are configuration flags of the same code path.

\subsection{Protocol and Metrics}
\emph{Learning phase}: 20 episodes, mission $\sim U\{$M1, M2$\}$, failure $\sim U\{$F1, F7$\}$, shared seeds across methods; SCBT over seeds 0--2, ablations over seed 0. \emph{Frozen phase}: crystallize; evaluate $T^\star_m$ on 20 curriculum and 20 held-out episodes with all learning disabled (no Patch Manager, farm, LLM, or bandit object is instantiated). B1 has no frozen artifact. \emph{Determinism}: 10 end-to-end repetitions of the first scenario (distinct trace hashes, tick spread), and for each repetition three offline re-ticks of the frozen tree against that repetition's recorded telemetry log with no-op commands (distinct hashes; whether the offline hash equals the online one). Metrics: success rate and mission value per failure type; abort rate; LLM calls per episode and cumulative (the \emph{amortization curve}); library hit rate (same/cross class); admission rejection rate per stage; recovery latency decomposed into LLM, tuning, and admission; bandit regret; parameter-transfer gap; crystallized-tree size versus B0; held-out generalization.

\section{Results}
\label{sec:results}

\subsection{Learning Phase}
\begin{figure}[t]
\centering
\includegraphics[width=0.95\columnwidth]{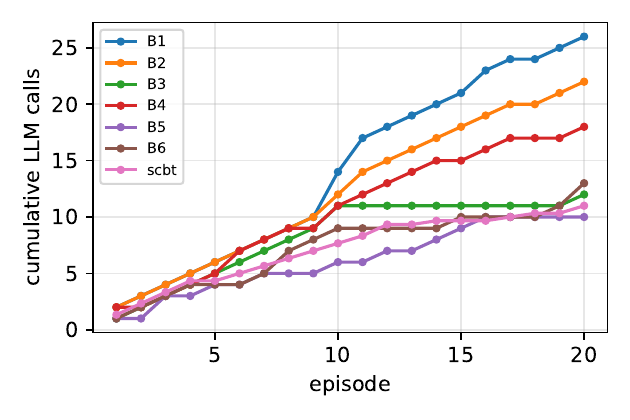}
\caption{Cumulative LLM calls versus episode (SCBT: mean over three seeds; others: seed 0). Methods without a library (B1, B2) remain near one call per episode; SCBT, B3, B5 and B6 flatten.}
\label{fig:amort}
\end{figure}

Table~\ref{tab:learn} and Fig.~\ref{fig:amort} summarize the learning phase. SCBT used $11\pm1.4$ LLM calls over 20 episodes and 0.33 calls per episode in the second half; whole-tree regeneration (B1) used 26 and patching without a library (B2) 22, both still above one call per episode at the end. The library is what produces amortization: B2 differs from SCBT only in not storing patches. The bandit's contribution is visible in B4, which calls the LLM whenever retrieval misses (18 calls, 0.70 per episode in the second half) but reaches the highest single-seed learning-phase success (0.80) by trying stored patches more often (25 hits). Across SCBT seeds, 12--14 of 23--24 failure events were served from the library.

Admission was selective: 2--3 proposals admitted and 7--9 rejected per seed, by stage~2 (landing reserve 0.6--8.0\%, or 238--1009 geofence-violation ticks under wind) and stage~3 (no margin over the current tree, nominal regression, or success below $\rho_{\min}$). Four rejections were admission \emph{errors} (malformed XML from the model reaching the compiler), which the pipeline treats as rejections. B5, which skips replays, admitted all 10 of its proposals. Per SCBT seed, admission consumed 103--203 replay episodes.

\begin{table*}[t]
\caption{Learning phase, 20 episodes, Gemini 3.1 Pro. SCBT: mean$\pm$sd over seeds 0--2; ablations: seed 0. ``Calls, ep.\ 11--20'' is LLM calls per episode over the second half. ``Hits'' are failure events served from the library.}
\centering\footnotesize
\begin{tabular}{lcccccccc}
\toprule
Method & Seeds & Success & Value & LLM calls & Calls/ep.\ 11--20 & Adm./rej. & Hits & Replays \\
\midrule
B1 regenerate whole tree & 1 & 0.55 & 0.50 & 26 & 1.20 & 11 / 11 & 0 & 308 \\
B2 no library              & 1 & 0.40 & 0.33 & 22 & 1.00 & 9 / 11  & 0 & 342 \\
B3 no CMA-ES tuning        & 1 & 0.65 & 0.57 & 12 & 0.10 & 3 / 9   & 10 & 120 \\
B4 no bandit (nearest nb.) & 1 & 0.80 & 0.65 & 18 & 0.70 & 4 / 10 & 25 & 252 \\
B5 static admission only   & 1 & 0.85 & 0.70 & 10 & 0.40 & 10 / 0  & 22 & 0 \\
B6 no cross-class transfer & 1 & 0.65 & 0.55 & 13 & 0.40 & 2 / 9   & 13 & 188 \\
\textbf{SCBT (full)}       & 3 & $0.75\pm0.14$ & $0.62\pm0.11$ & $11\pm1.4$ & 0.33 & 2.3 / 8.0 & 12.3 & 103--203 \\
\bottomrule
\end{tabular}
\label{tab:learn}
\end{table*}

\subsection{Frozen Phase}
Table~\ref{tab:frozen} reports the crystallized trees evaluated with every learning component disabled. SCBT's trees, with 28 nodes for M1 and 33 for M2 and two patches each, reached $97\pm2\%$ success and $0.78\pm0.04$ mission value on the curriculum, against 12\% and 0.12 for the unpatched base tree and 98\%/0.83 for the hand-coded expert tree B0. Per failure type, SCBT recovered 22/24 F1 and 36/36 F7 episodes (B0: 23/24 and 36/36). A tree no human wrote matches, on the distribution it learned from, one written by an engineer with the failure catalogue in hand, with no model in the loop.

On held-out variants every method degrades: F1 held-out gusts (28--30\,m/s for 30--36\,s) push the vehicle outside the geofence in 14 of 20 episodes regardless of the tree, and B0 recovers only 4/36 of them. SCBT (4/36 F1, 22/24 F7) is statistically indistinguishable from B0 there. The exception is B5, discussed below.

\begin{table*}[t]
\caption{Frozen phase: crystallized trees with all learning disabled, 20 curriculum and 20 held-out episodes per seed (B0, base, SCBT: seeds 0--2; ablations: seed 0). ``End-to-end'': distinct trace hashes over 10 repetitions of one scenario (worst seed) and maximum tick spread. ``Offline'': distinct hashes over 3 re-ticks of the frozen tree against each repetition's recorded telemetry, and the number of logs whose offline hash equals the online one. Nodes: M1/M2 trees (patches applied).}
\centering\footnotesize
\begin{tabular}{lcccccccc}
\toprule
 & \multicolumn{2}{c}{Curriculum} & \multicolumn{2}{c}{Held-out} & \multicolumn{2}{c}{End-to-end} & Offline & Nodes \\
\cmidrule(lr){2-3}\cmidrule(lr){4-5}\cmidrule(lr){6-7}
Method & Succ. & Value & Succ. & Value & Distinct & Spread & Dist./faithful & M1 / M2 \\
\midrule
Base (unpatched)  & $.12\pm.09$ & $.12\pm.09$ & 0.00 & 0.00 & -- & -- & -- & 15 / 20 (0) \\
B0 expert & $.98\pm.02$ & $.83\pm.06$ & $.47\pm.06$ & $.37\pm.08$ & 1 / 10 & 0 & 1/3, 30/30 & -- \\
B2 no library          & 1.00 & 0.85 & 0.40 & 0.30 & 1 / 10 & 0  & 1/3, 10/10 & 77 / 82 (9+9) \\
B3 no CMA-ES tuning    & 1.00 & 0.85 & 0.45 & 0.35 & 4 / 10 & 33 & 1/3, 10/10 & 33 / 38 (3+3) \\
B4 no bandit           & 0.80 & 0.62 & 0.15 & 0.10 & 1 / 10 & 0  & 1/3, 10/10 & 37 / 42 (4+4) \\
B5 static adm.\ only & 0.95 & 0.80 & \textbf{1.00} & \textbf{0.75} & 1 / 10 & 0 & 1/3, 10/10 & 84 / 89 (10+10) \\
B6 no transfer & 1.00 & 0.82 & 0.45 & 0.30 & 2 / 10 & 30 & 1/3, 10/10 & 33 / 38 (2+2) \\
\textbf{SCBT (full)} & $\mathbf{.97\pm.02}$ & $.78\pm.04$ & $.43\pm.10$ & $.32\pm.10$ & 3 / 10 & 12 & 1/3, 30/30 & 28 / 33 (2+2) \\
\bottomrule
\end{tabular}
\label{tab:frozen}
\end{table*}

\subsection{Determinism}
In all eleven frozen cells, re-ticking the crystallized tree offline against a recorded telemetry log produced a single trace hash over three re-ticks, and that hash equaled the hash of the online run that produced the log (60 of 60 logs across all cells). The control logic is therefore a pure function of its observations, as claimed. End to end, seven cells gave one hash in ten repetitions; the remaining four (B3, B6, two SCBT seeds) gave 2--4 distinct hashes with tick spreads of 1--33 ticks (0.05--1.65\,s) and the same final outcomes. We traced the residual to the simulator: its command-execution thread is scheduled by wall-clock time, so lockstep stepping is exact (a 50\,ms request always advances 17 physics steps) but the moment a command takes effect within a step is not, and trajectories drift at the $10^{-4}$\,m level. Threshold crossings (arrival tolerance, guard conditions) occasionally flip by one tick. We report this rather than hide it, since a reader deploying crystallized trees on hardware will face the same distinction: the tree is deterministic, the world is not.

\subsection{What Admission Caught, and What It Got Wrong}
\label{sec:caught}
Three observations argue for empirical admission; a fourth argues against the particular stage-2 rule we used.
\begin{enumerate}
\item \textbf{Safe but useless patches.} With static-only admission, the model's first battery recovery was \texttt{ReturnToLaunch} alone: safe, admitted, and 0 for 6 on mission success because it abandoned the required goal. The bandit then correctly concluded the LLM was not worth $\kappa$ and converged on the bad patch. Stage~3 rejects such patches for lack of margin.
\item \textbf{Optimizer exploiting the objective.} CMA-ES without the safety term chose a 13.2\% reserve that landed at 9.8\%; adding the invariant penalty to (\ref{eq:J}) yielded 13.9\% and 12.5\% at landing. Across the matrix, stage~2 rejected five battery patches that would have landed at 0.6--8.0\%.
\item \textbf{Nominal regression.} Before the nominal-replay check, a \textsc{ReplaceSubtree} that deleted the optional survey lanes was admitted, used seven times with seven successes, and crystallized: it never failed a required goal because it never attempted an optional one. The check now rejects any drop in nominal mission value.
\item \textbf{Blaming the patch for the wind.} Stage~2 rejects a patch if \emph{any} replay violates an invariant, without reference to what the unpatched tree does under the same perturbation. Under F1 the gust itself carries the vehicle across the geofence; replays of a wind-recovery patch therefore logged 238--1009 violation ticks and the patch was rejected, although the violation was not of its making. SCBT consequently crystallized with few wind recoveries, whereas B5, which admits every statically valid patch, kept eight \textsc{WrapFallback} and two \textsc{WrapRetry} patches (84/89 nodes) and recovered 14/14 held-out F1 episodes that no other method, including the expert tree, could. B5's frozen trees were not unsafe on this curriculum (no invariant violations in 20 curriculum episodes; three landings below 10\% reserve, versus two for SCBT). The correct stage-2 rule is \emph{relative}: reject violations the patch introduces beyond those of the unpatched tree under the same seeds, as stage~3 already does for success. We leave this as the first change for future work rather than re-run with it, so that the reported numbers reflect the pipeline as specified in Sec.~\ref{sec:admission}.
\end{enumerate}

\section{Discussion}
\label{sec:discussion}

\textbf{What crystallization buys.} The frozen artifact is deterministic and inspectable, and provenance ties every subtree to the failure that produced it, the model's rationale, and the admission record. It does not generalize beyond what the learning phase encountered; the held-out metric exists to make that limit visible.

\textbf{Patches versus regeneration.} Whole-tree regeneration (B1) discards verified logic, makes differences between episodes unattributable, and in our runs used the most LLM calls and the most replays (308) while achieving the lowest learning-phase success among library methods. Patches preserve attribution and compose; the crystallized SCBT tree is the base tree plus two readable edits.

\textbf{Admission is a dial, not a filter.} The B5 result is the most informative in the matrix: on this curriculum, admitting every statically valid patch produced larger trees that generalized better, and the safety cost was small. Our stage-2 rule was too strict in one specific, correctable way (Sec.~\ref{sec:caught}); but the general point stands that admission trades coverage for guarantees, and the right setting depends on how much the operator trusts the simulator's invariants versus the model's proposals. The framework makes that trade explicit and measurable; it does not make it for you.

\textbf{Safety.} Admission is necessary, not sufficient. Randomized replay around a recorded event does not cover the deployment distribution; the invariant monitors are the operative safety mechanism and run identically in replay and in the live episode. We make no formal-verification claim.

\textbf{Limitations.} Two failure types; single vehicle per episode (the fleet, F12 and cross-class transfer are implemented but not exercised, so B6 differs from SCBT only by configuration); ablations on one seed; headless host, so context is telemetry-derived and the perception-dependent failures are absent; Simple Flight rather than PX4. The held-out F1 band is largely infeasible for every method and therefore measures robustness to an unrecoverable condition rather than generalization; a held-out band inside the recoverable regime would be more informative. Four admission errors were caused by malformed model output reaching the compiler and should be caught at the static stage.

\section{Conclusion}
\label{sec:conclusion}
The adaptability of foundation models and the determinism of classical control are usually presented as a choice. SCBT sequences them: an LLM proposes structure when a Behavior Tree fails, CMA-ES tunes it, replay admission gates it, a bandit learns when to stop asking, and the library compiles into an ordinary deterministic tree. On Project AirSim, the learning phase needed $11\pm1.4$ LLM calls in 20 episodes against 26 for regeneration, and the crystallized trees, with two patches each, matched a hand-coded expert tree on the training distribution with no model running and with decision traces that are an exact function of their observations. The admission pipeline caught safe-but-useless patches, an optimizer exploiting its objective, and a patch that deleted mission goals; it also wrongly rejected wind recoveries by attributing the wind's geofence excursions to the patch, and the static-only ablation generalized better as a result. Future work: relative invariant checks in admission, the fleet and cross-class experiments, the rendering host with VLM context, and deployment of crystallized trees on PX4 hardware.

\section*{Acknowledgment}
The author thanks the IAMAI Simulations team for maintaining Project AirSim and its headless runtime, on which all experiments were run.

\end{document}